\documentclass{article}

\usepackage{arxiv}
\usepackage[utf8]{inputenc} % allow utf-8 input
\usepackage[T1]{fontenc}    % use 8-bit T1 fonts
\usepackage[colorlinks=true,linkcolor=blue, citecolor=blue]{hyperref}       % hyperlinks
\usepackage{url}            % simple URL typesetting
\usepackage{booktabs}       % professional-quality tables
\usepackage{amsfonts}       % blackboard math symbols
\usepackage{nicefrac}       % compact symbols for 1/2, etc.
\usepackage{microtype}      % microtypography
\usepackage{lipsum}		% Can be removed after putting your text content
\usepackage{graphicx}
\usepackage{doi}
\usepackage{wrapfig}
\usepackage{subcaption}
\usepackage{xcolor}

\graphicspath{ {./images/} }

\title{Identical Image Retrieval using Deep Learning}

\author{Sayan Nath \\
	School of Computer Engineering\\
	Kalinga Institute of Industrial Technology\\
	India \\
	\texttt{1906426@kiit.ac.in} \\
	\And
	Nikhil Nayak \\
	School of Computer Engineering\\
	Kalinga Institute of Industrial Technology\\
	India \\
	\texttt{1928239@kiit.ac.in} \\
}

\date{}

\renewcommand{\headeright}{}
\renewcommand{\undertitle}{}

\hypersetup{
pdftitle={Identical Image Retrieval using Deep Learning},
pdfsubject={Image Search},
pdfauthor={Sayan Nath, Nikhil Nayak},
pdfkeywords={Image Similarity Search, Computer Vision, Deep Learning, Transfer Learning, BigTransfer},
}

\begin{document}
\maketitle

\begin{abstract}
In recent years, we know that the interaction with images has increased. Image similarity involves fetching similar-looking images abiding by a given reference image. The target is to find out whether the image searched as a query can result in similar pictures. We are using the BigTransfer Model, which is a state-of-art model itself. BigTransfer(BiT) is essentially a ResNet but pre-trained on a larger dataset like ImageNet and ImageNet-21k with additional modifications. Using the fine-tuned pre-trained Convolution Neural Network Model, we extract the key features and train on the K-Nearest Neighbor model to obtain the nearest neighbor. The application of our model is to find similar images, which are hard to achieve through text queries within a low inference time. We analyse the benchmark of our model based on this application.
\end{abstract}

% keywords can be removed

\section{Introduction}
How can we compute the similarity of one image to another? This is a question that has been asked for hundreds of years, and it is probably the most fundamental question in image processing. Several customer-facing applications leverage images to search and find products and they usually complement that of a text-based search in most use cases. In this paper, we will describe an approach to compute image similarity using deep neural networks. Our method is based on a state-of-the-art model known as the BigTransfer Model, which learns to predict the similarity of two images. Fig 1. Shows examples of Image samples from the tf\_flower \cite{dataset} dataset on which our BigTransfer(BiT) \cite{BiT} model is trained. 

\begin{figure}[h]
\includegraphics[width=7.5cm, height=7cm]{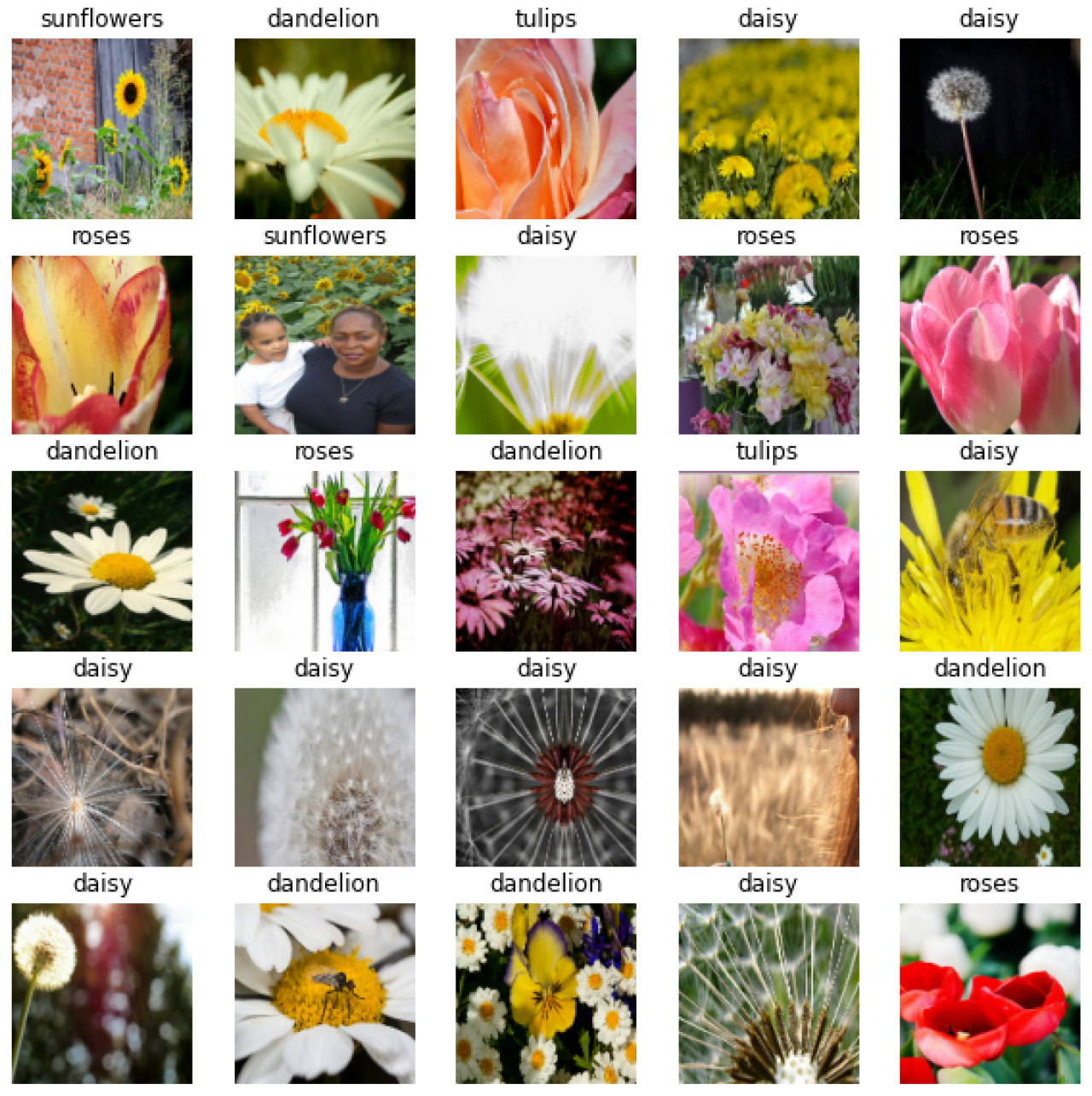}
\centering
\caption{TensorFlow Flower Dataset Sample}
\end{figure}

Our system has the potential to be used with business-critical applications. With the visual exploration trend rising in the retail sector and the availability of quintillion Gigabytes of data at our disposal our image similarity model is bound to become more and more accurate. We optimized our model for two specific objectives: search accuracy and query duration. We achieved high accuracy by examining a few other candidates. We achieved a query time of under a second. Our system is highly accurate and fast. Our system is fast because we used a few other models such as hand-crafted features, autoencoders \cite{autoencoders}, CNN's \cite{CNN} and we only used the most recent pre-trained model.

This approach is highly robust and we believe it could be adopted by the industry. We chose a specific type of deep learning model because of the best performance we have observed and because of the clarity of the approach. The state-of-the-art model we chose, based on the results, can leverage unsupervised learning. With that type of model, it is easier to train data in massive quantities and to have a much better generalization. BigTransfer(BiT) \cite{BiT} model uses a feed-forward layer over an unsupervised autoencoder \cite{autoencoders} along with an attention model \cite{Attention}. Our paper reports the first use of an attention model \cite{Attention} for model-based image retrieval. Model-based image retrieval has been used for a long time in text-based searches for making more accurate queries. But image-based search and image retrieval are very different scenarios. Visual perception works differently than text-based search. So is image-based search and image retrieval.  Many other applications besides those highlighted in this paper can be benefitted from this work. Our system is robust, highly accurate, and it can leverage unsupervised learning to produce large amounts of high-quality training data. This has enormous potential in several domains.

While image search engines like Google or Bing can find similar images with a text search, they are not very effective for a variety of reasons. The most obvious reason is that textual descriptions are limited to 20,000 characters and cannot express detailed semantics or the history of an object. Secondly, people who don’t speak English cannot query image search engines via language translation. This brings about the second problem of a large and growing number of users from emerging markets who may not have access to Google and Bing. More importantly, they do not have a search query language of their own, yet still, they express their need to find images of stars and moons using such complex terms that require a series of complex symbols or gestures to communicate. Indeed, the search relevance of image search engines is heavily dependent on the existence of similar images available on the Internet. This is an issue for all images that are not publically accessible for use, images captured at different times, or cameras with varying parameters. This is a very important point that we take into consideration when working on this project. On one side we have images that are not publicly accessible. The quality of the images will be varied and they will require careful content selection in the selection and processing of training and test data. On the other side, there is a lot of commercial information that "needs to be made publicly available for commercial use but is not yet. It can be difficult for a company to get permission to use their data, and another difficult process to get the images in the public domain. We also expect to expand and improve our existing models to make the selected publicly available images accessible for future work. However, in this project, we are primarily focusing on developing our data collection pipeline to collect images that are currently out of the public domain and publish them on our existing system. The project's goal is to assist small and medium enterprises to more easily produce a high-quality dataset in the image and document domains. The paper has used all possible efforts to make our project as replicable and reproducible as possible.

So far we have classified on flower dataset \cite{dataset} and we have employed an unsupervised feature extraction technique for image preprocessing. In the future, we will release the new datasets, which can be useful to anyone else for training and deploying their system for image retrieval. Also in the future, we will explore a method to integrate our images of objects into deep search engines to improve the depth of the search results. This will add some search value, which is useful to some specific applications such as content discovery on an exhibition. We expect that this will drive towards some additional use cases in the future. Research and industry have benefited enormously from advances in digital communication technology. This technology has enabled tremendous growth in markets such as mobile commerce, social networking, and large-scale community projects. In this project, we envisage that such image-based platforms, which focus on image-based search \cite{SIR, deepcnn}, discovery, and mobile e-commerce, will present a massive opportunity for new business models.

\section{Related Work}
\label{sec:headings}

\subsection{Similar Image Retrieval (SIR)}

The central idea is to convert each image into a fingerprint, signature, or unique descriptor \cite{SIR}. Internally, fingerprints are basically embeddings calculated from a suitable deep neural network. We have tried many generations of embedding techniques and adopted VGG16 \cite{VGG} as our primary network. The embedding is removed from the last fully connected layer of VGG16 \cite{VGG}. A typical embedding is a high-dimensional vector consisting of floating-point numbers. In the binarized format, the embedding is broken down into smaller subcodes that are included in the Elasticsearch index. When searching, we also use Elasticsearch's ability to perform efficient Hamming distance calculations in the form of bit-wise operations. Business applications are primarily focused on images that have been added to the catalogue in the last few weeks. Therefore, design the indexing process as a rolling process to constantly index new and recently updated images. The currently deployed system listens for Kafka themes that stream new and updated images. The running index (last 3 months) of the newly created image is maintained for later search and retrieval \cite{SIR}. Due to the continuous nature of the application, hash-based indexing is the preferred choice over the technique of collectively learning representations from static datasets such as Principal component analysis (PCA) \cite{PCA}. As the catalogue changes, so do the best key components, which require frequent recalculations. During the search phase, the query image (also known as the seed image) is provided to the system via the front end. On the backend, the query image is converted to an embed and its closest neighbour \cite{KNN} is retrieved from the indexed store. The retrieved images are displayed in the grid in descending order of similarity to the query. Each resulting image has a checkbox that allows the user to select only the relevant image from the grid.

\subsection{Image similarity with Deep CNN}

Image validation algorithms aim to determine if a particular pair of images is similar \cite{deepcnn}. Image verification is not similar to image identification. The former solves similar image use cases, while the latter is due to the nature of image retrieval. Advances in image verification relate to two main areas: image embedding and metric learning \cite{metric}. Image embedding trains robust and discriminating descriptors to represent each image as a compact feature vector/embedding. The current state-of-the-art function descriptor is generated by the self-learning function CNN \cite{CNN}. SimNet \cite{SimNet} uses multiscale CNN \cite{Multi-scale-CNN} in a Siamese network \cite{Siamese} that learns 4096-dimensional embedding of images. Pairing is required for the Siamese network \cite{Siamese}. A pair of positive images (almost similar images) and a pair of negative images (different images). This allows you to learn the range of intervals. It turns out that choosing the right image pair for training is very important for achieving good model performance and faster model convergence. We propose a new online pair mining (OPMS) strategy that attempts to steadily increase the difficulty of image pairs during network training. Multiscale CNN \cite{Multi-scale-CNN} is used in  Siamese networks. This CNN \cite{CNN} learns common image embeddings in the upper and lower layers. This model learns much better image embedding than traditional CNNs \cite{CNN} for image similarity tasks.

\section{Proposed Method}
\label{sec:headings}

We have used BigTransfer(BiT) \cite{BiT} as our pre-trained CNN model \cite{CNN} to extract the features. BigTransfer(BiT) \cite{BiT} is a set of pre-trained image models that can be transferred to obtain excellent performance on newer datasets, even with few examples per class. The performance of the BigTransfer(BiT) \cite{BiT} model increases as the dataset size increases. Transferring of pre-trained representations improves the efficiency and simplifies hyperparameter tuning when we train a deep neural network for computer vision. BigTransfer(BiT) \cite{BiT} is trained in a large, generic dataset and its weights are then used to initialise subsequent tasks that can be solved with fewer data points, and less computation. BigTransfer(BiT) \cite{BiT} is trained on three large datasets i.e JFT-300M dataset \cite{JFT}, which contains 300M noise labelled images. BigTransfer(BiT) \cite{BiT} is transferred to execute many diverse tasks by setting the train set sizes from one example per class to one million examples per class. BigTransfer(BiT) \cite{BiT} is highly effective and provides insight into the interplay between scale, architecture and training hyperparameters.

\begin{figure}[h]
\includegraphics[width=9.5cm]{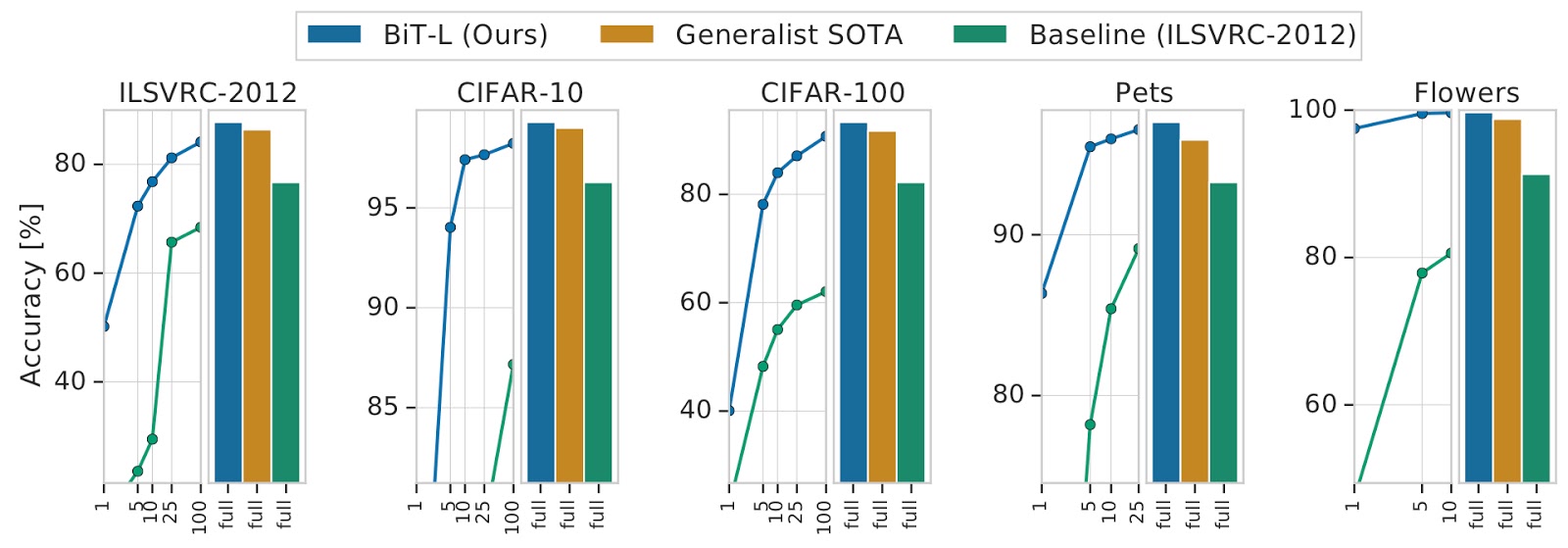}
\centering
\caption{The x-axis shows the number of images used per class, ranging from 1 to the full dataset. On the plots on the left, the curve in blue above is our BiT-L model, whereas the curve below is a ResNet-50 pre-trained on ImageNet (ILSVRC-2012)}
\end{figure}

We are extracting the BigTransfer(BiT) \cite{BiT} model to extract the image features and using those features we are calculating the distance between the images using the reference image. The greater the distance the more likely the images are alike. It scales well to millions of data points, and It achieves a good trade-off between the probability of error and the number of data points needed to achieve this probability.

\begin{figure}[h]
\includegraphics[width=\linewidth]{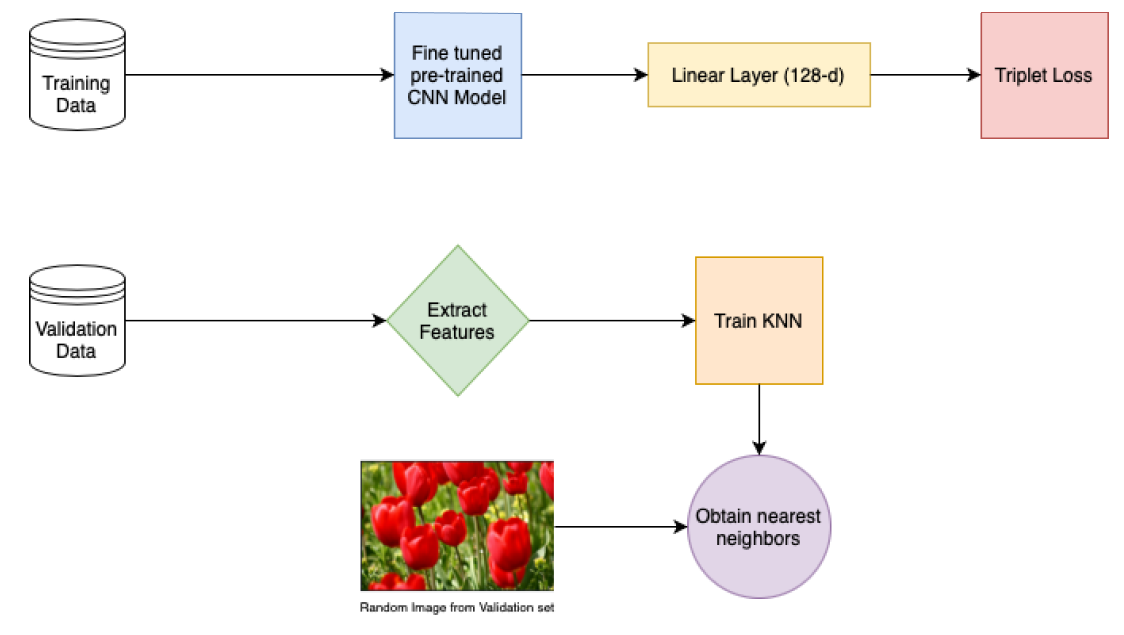}
\centering
\caption{Proposed Architecture}
\end{figure}

\section{Experiments}
\label{sec:headings}

\subsection{Experimental Setup}

In this experiment, we have used the dataset of flowers \cite{dataset} provided by tensorflow. It is popularly known as \texttt{tf\_flower} \cite{dataset}. TensorFlow Flower Dataset \cite{dataset} consists of five classes. Five classes are labelled as 'Daisy', 'Dandelion', 'Roses', 'Sunflowers', 'Tulips'.  The number of classes for the dataset is imbalanced.

\begin{figure}[h]
\includegraphics[width=7.5cm]{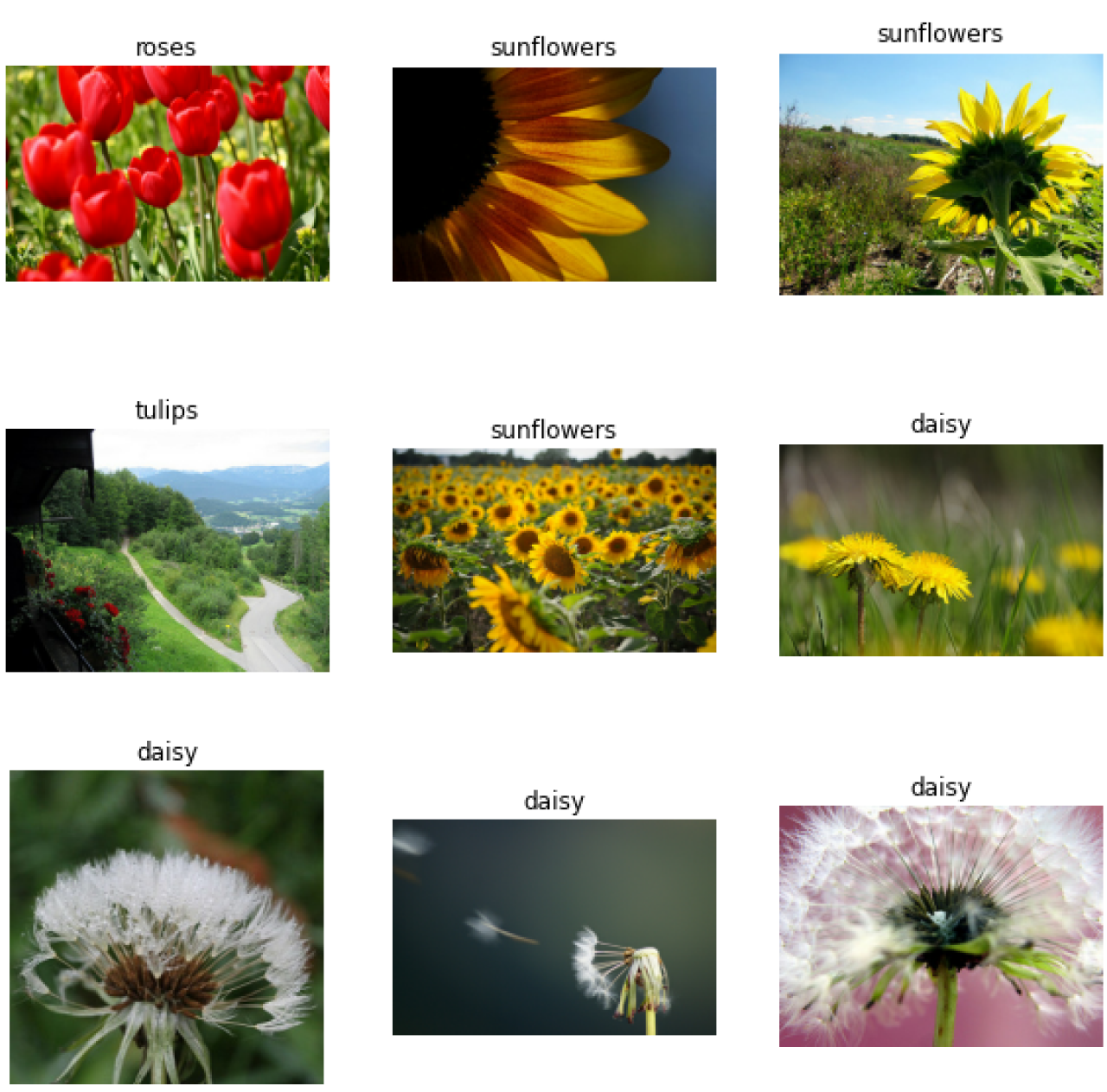}
\centering
\caption{Visualization of TensorFlow Dataset before training}
\end{figure}

The dataset is being loaded using the tensorflow dataset \cite{dataset}. The dataset is then divided into train and validation data. 85\% of the data is taken as the training data and the rest of the data is taken as validation data. The number of training and validation samples were 3120 and 550. To make our model more robust I applied data augmentation \cite{augment, augmentation}. Used random flip with horizontal and vertically on the images. Used a random rotation with a factor of 0.2. Applied random zoom at a height and width factor of 0.2. Training images are resized to 160 and images are randomly cropped to 128. Validation images are resized to 160. 

We loaded the pre-trained BigTransfer(BiT) \cite{BiT} model which is trained on ImageNet21k \cite{ImageNet21k} downloaded from TensorFlow Hub \cite{tfhub, tfhub-biT, bit-github}. We created a BigTransfer(BiT) \cite{BiT} model and normalised the dense representation. Used TripletSemiHardLoss \cite{tripletloss} as a loss function. The loss encourages the positive distances (between a pair of embeddings with the same labels) to be smaller than the minimum negative distance among which are at least greater than the positive distance plus the margin constant (called semi-hard negative) in the mini-batch. If no such negative exists, use the largest negative distance instead. 

\begin{equation}
	\left \| f(x_{i}^{a}) - f(x_{i}^{p}) \right \|_{2}^{2}\textrm{} < \left \| f(x_{i}^{a}) - f(x_{i}^{n}) \right \|_{2}^{2}\textrm{}
\end{equation}

We used Scholastic Gradient Descent \cite{gradient-descent, stochastic-gradient-descent, stochastic-gradient-descent-two} as our optimizer with a variable learning rate and momentum of 0.9. The learning rate was decaying by a factor of 10 at schedule boundaries.

 \begin{center}
 for i in range (m): 
 \end{center}
 
\begin{equation}
	\Theta {j} = \Theta {j} - \alpha \left ( \widehat{y}^{i} - y^{i} \right )x_{j}^{i}
\end{equation}

After compiling the model and the callbacks for the models are set up. Used Early Stopping to monitor the validation loss with a patience rate of 5 and standard rate for verbose i.e 2. Used CSV Logger to log all the data during training the model. 

\subsection{Experimental Results}

After setting up the experiment, we trained our BigTransfer(BiT) \cite{BiT} model on the flower dataset \cite{dataset}. Our training started with a loss of 0.94 and a validation loss of 0.91. The training is being done on Tesla V100-SXM2. The average time for an epoch is 2 seconds. The training time was 78.68 seconds with only 21 epochs. Our loss and validation loss after 21 epochs is 0.23 and 0.30 respectively. 

\begin{figure}[h]
    \centering
    \begin{subfigure}[b]{0.3\textwidth}
        \includegraphics[width=\textwidth]{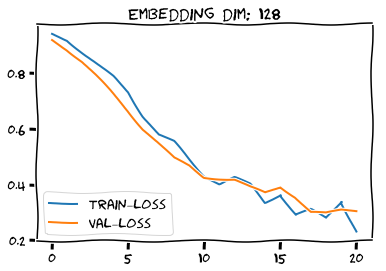}
        \caption{Training Graph}
        \label{fig:gull}
    \end{subfigure}
    \begin{subfigure}[b]{0.3\textwidth}
        \includegraphics[width=\textwidth]{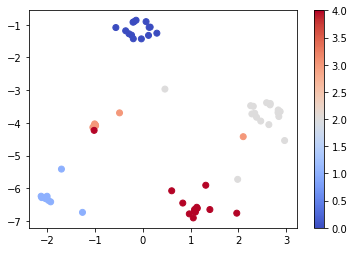}
        \caption{Visualizing the Embedding Space for Current Validation Batch}
        \label{fig:gull}
    \end{subfigure}
    \begin{subfigure}[b]{0.3\textwidth}
        \includegraphics[width=\textwidth]{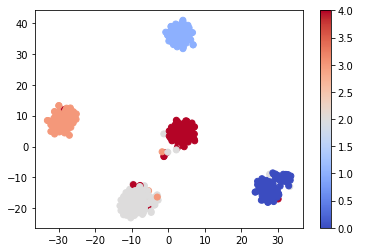}
        \caption{Visualizing the Embedding Space for Entire Validation Batch}
        \label{fig:tiger}
    \end{subfigure}
    \caption{Visualizing the Results}\label{fig:animals}
\end{figure}

Saved the BigTransfer(BiT) \cite{BiT} model after training the model on flower dataset \cite{dataset}. Defining the BigTransfer(BiT) Model to load the model weights we saved as our Keras model. Created a validation pipeline for training the nearest neighbor \cite{KNN} model. We are calculating our nearest neighbor \cite{KNN} for the features of our query image. The model took 0.00043 seconds for 550 samples.  

\begin{figure}[h]
\includegraphics[width=15cm, height=18.5cm]{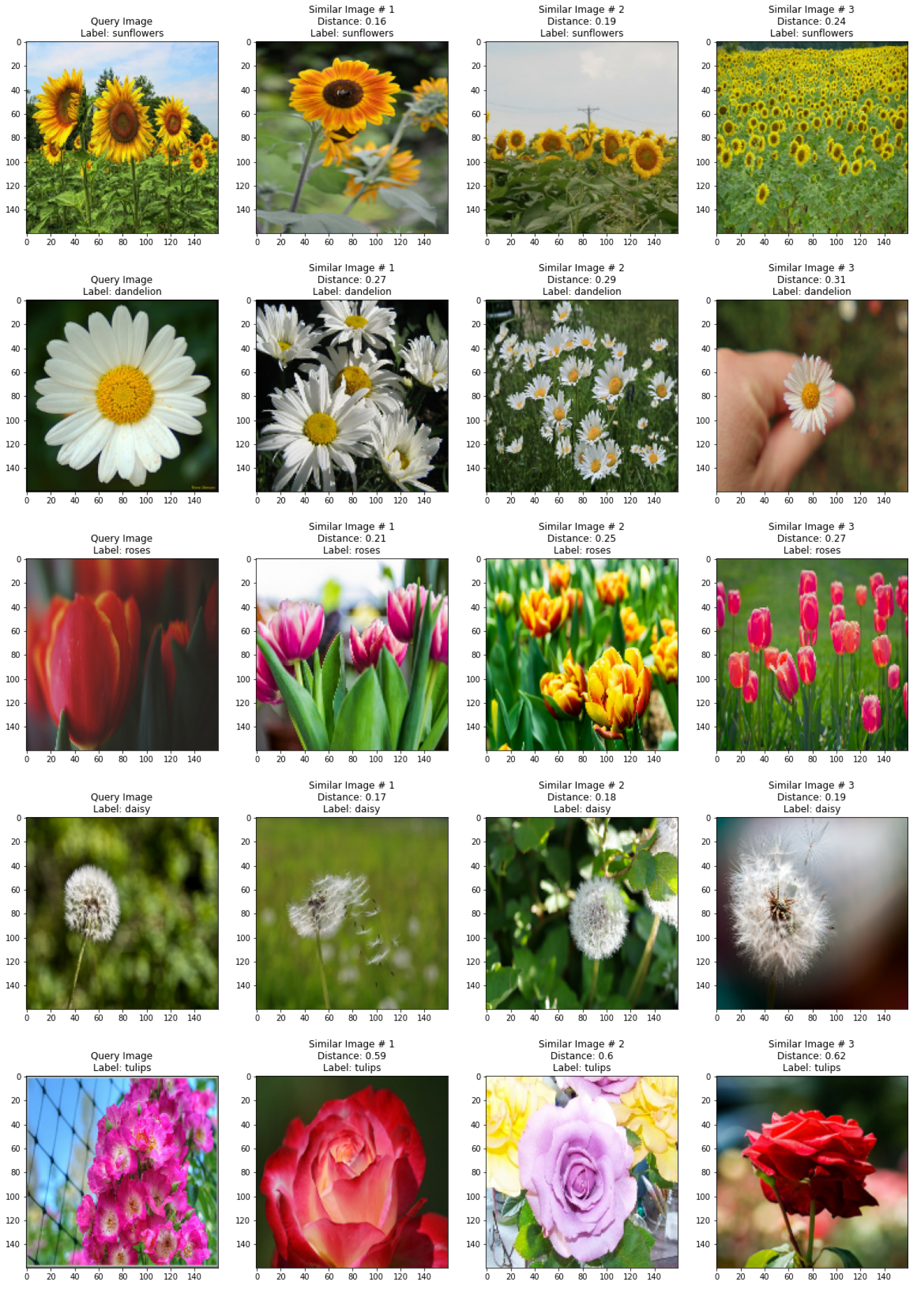}
\centering
\caption{Results of the Experimentation}
\end{figure}

\section{Conclusion}
\label{sec:headings}

This project shows how to scale up the pre-trained model by training on a larger dataset to extract key features. The model is based on a Resnet152 \cite{Resnet} backbone and the classifier is an FCN-8s. The model is based on the code release for BiT-M. BigTransfer(BiT) \cite{BiT} does not require any pre-processing to the input images, i.e., we can use vanilla ImageNet models to transfer on any visual tasks. We see this as an important step towards practical implementations. The idea behind our model is to build highly accurate and low latency visual search tools that can fit in with many business facing applications.

\end{document}